\documentclass[conference]{IEEEtran}
\IEEEoverridecommandlockouts
\usepackage{cite}
\usepackage{amsmath,amssymb,amsfonts}
\usepackage{algorithmic}
\usepackage{graphicx}
\usepackage{textcomp}
\usepackage{multirow}
\usepackage{booktabs}
\usepackage[table]{xcolor}
\usepackage{array}
\usepackage{makecell}
\usepackage{colortbl} 
\def\BibTeX{{\rm B\kern-.05em{\sc i\kern-.025em b}\kern-.08em
    T\kern-.1667em\lower.7ex\hbox{E}\kern-.125emX}}
\begin{document}

\title{A Text-to-Game Engine for UGC-Based Role-Playing Games\\}

\author{\IEEEauthorblockN{Lei Zhang}
\IEEEauthorblockA{\textit{RPGGO} \\
codingtmd.eth@rpggo.ai}
\and
\IEEEauthorblockN{Xuezheng Peng}
\IEEEauthorblockA{\textit{RPGGO} \\
pengxzh@rpggo.ai}
\and
\IEEEauthorblockN{Shuyi Yang}
\IEEEauthorblockA{\textit{RPGGO} \\
\textit{pptbt29@rpggo.ai}}
\and
\IEEEauthorblockN{Feiyang Wang}
\IEEEauthorblockA{\textit{RPGGO} \\
feiyang@rpggo.ai}
}

\maketitle

\begin{abstract}
The transition from professionally generated content (PGC) to user-generated content (UGC) has reshaped various media formats, encompassing formats such as text and video. With rapid advancements in generative AI, a similar transformation is set to redefine the gaming industry, particularly within the domain of role-playing games (RPGs). This paper introduces a novel framework for a text-to-game engine that leverages foundation models to transform simple textual inputs into intricate, multi-modal RPG experiences. The engine dynamically generates game narratives, integrating text, visuals, and mechanics, while adapting characters, environments, and gameplay in real time based on player interactions. To evaluate and demonstrate the feasibility and versatility of this framework, we developed the ‘Zagii’ game engine. Zagii has successfully powered hundreds of RPG games across diverse genres and facilitated tens of thousands of online gameplay sessions, showcasing its scalability and adaptability. These results highlight the framework’s effectiveness and its potential to foster a more open and democratized approach to game development. Our work underscores the transformative role of generative AI in reshaping the gaming lifecycle and advancing the boundaries of interactive entertainment.

\end{abstract}


\section{Introduction}
The creation of traditional RPGs, typically undertaken by professional and large development teams, requires a diverse set of skills. These include screenwriting, character design, game mechanics, coding, and graphical design, among others. This multidisciplinary requirement makes the development process costly, time-consuming, and relatively inflexible, as changes in one area often trigger cascading changes across the game. This complexity slows the introduction of new games to the market, restricts the variety of available narrative and gameplay styles, limits player freedom and control over narrative paths, and affects content expansion depth. Traditional RPGs, reliant on rigid game engines, often struggle to dynamically adapt to player choices, typically offering a linear or branching path that doesn’t evolve based on player interaction, or networked narratives that aren’t replicable for broader use.

\begin{figure}[t] 
    \centering 
    \includegraphics[width=0.48\textwidth]{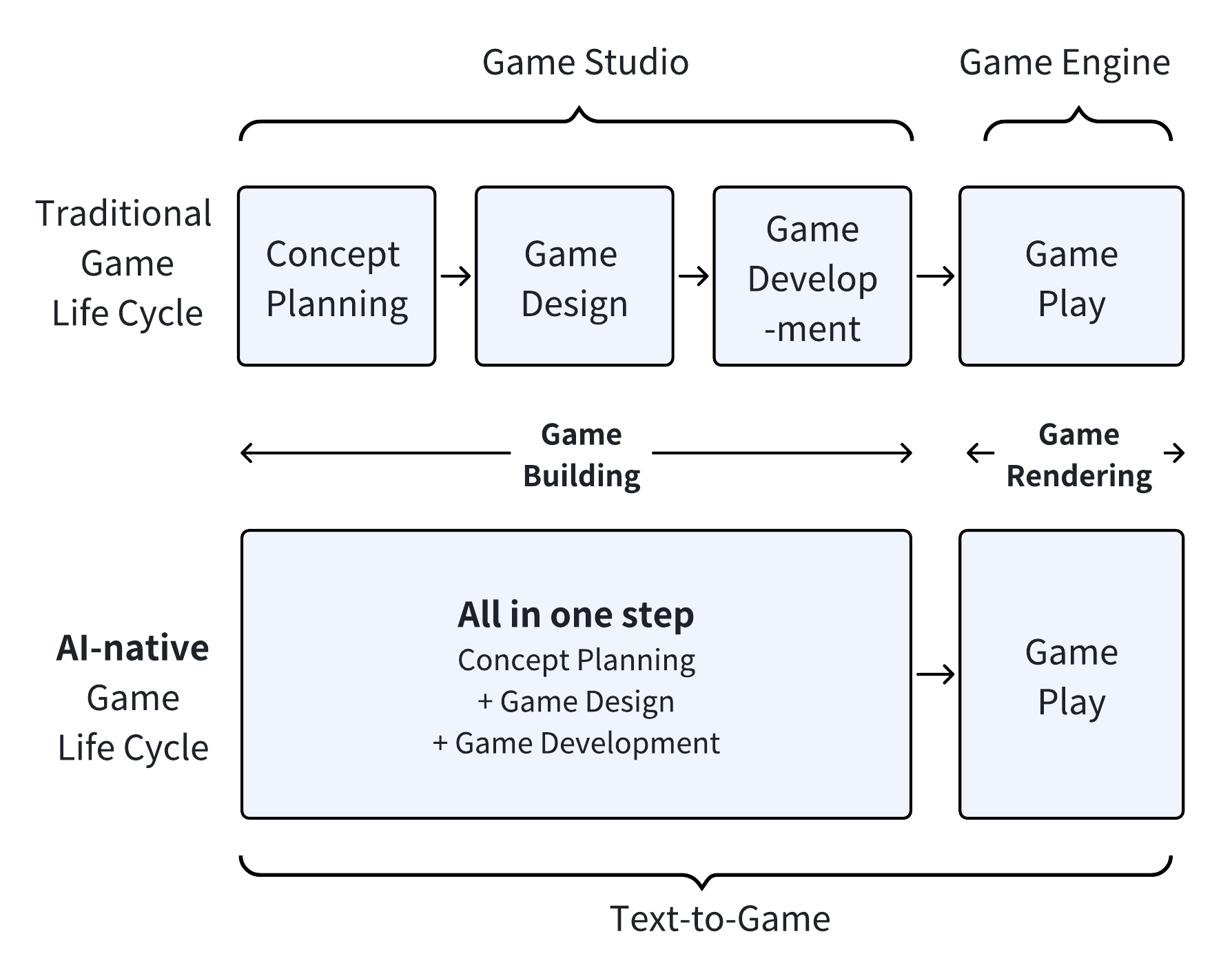} 
    \caption{The game life cycle difference between traditional RPGs and AI-native RPGs} 
    \label{fig:text-to-game} 
\end{figure}

Fig.~\ref{fig:text-to-game} illustrates the major life cycle of a game, structured into several phases: Concept Planning, Game Design, Game Development, and Game Rendering. Developers meticulously handcraft these phases, requiring substantial manual effort and creativity. Asset generation involves professional artists and designers creating high-quality characters, environments, and props. Despite the detailed craftsmanship, the user experience often features an accumulation of artificial elements, limited game paths, and finite experiences, resulting in a predictable and repetitive set of dialogues, endings, and player interactions.

In contrast, AI-native RPGs represent a significant paradigm shift in game development. These games utilize generative AI to evolve, generate, and maintain game content from a certain initial condition, thereby eliminating the need for human intervention. This AI-driven approach allows for the synthesis of various game elements, such as storylines, characters, and worlds, from simple textual inputs provided by users. Unlike traditional methods, this approach significantly reduces the need for technical skills, enabling individual creators to produce complex games. Furthermore, AI-native RPGs are inherently dynamic, with the unique capability to dynamically generate and adjust game content in real-time based on player decisions. This adaptability results in a more personalized gaming experience, as the game world evolves uniquely for each player, reflecting their actions and choices in a narrative that continuously unfolds.

In this paper, we present a practical development process for AI-native RPGs and propose a Generative AI-based Text-to-Game engine. Additionally, we delve into the design principles and technical challenges associated with the core modules of this AI game engine. Through the successful implementation of the complete development workflow, we have facilitated the creation of hundreds of games and supported tens of thousands of gameplay sessions. These results validate the feasibility of our approach and highlight the immense potential of AI-native RPGs as a promising and valuable area for future research.

\section{Related Works}

There are numerous works at the intersection of Large Language Models (LLMs) and game development, the role of LLMs in games can be categorized into several key areas: Player, Non-Player Character (NPC), Game Master (GM), Player Assistant, Commentator/Reteller, Game Mechanic, Automated Designer, and Design Assistant\cite{gallotta2024large}. LLMs can play games by converting game states and actions into token sequences, handling both text-based and visual-based game states. They have been applied in board games like Chess\cite{toshniwal2022chess}, Go\cite{ciolino2020go}, and Othello\cite{li2022emergent}, as well as in text adventure games where they generate character response based on environment descriptions\cite{yao2020keep}\cite{tsai2023can}. LLMs also play Atari games by predicting actions from visual inputs, as demonstrated by the GATO agent\cite{reed2022generalist}.

In enhancing NPC dialogue and behavior, LLMs create immersive interactions by adapting responses to game contexts\cite{shanahan2023role}. They are used for both foreground NPCs, which require contextual interactions\cite{warpefelt2017model}\cite{xu2023exploring}\cite{maas2023infinity}, and background NPCs, which maintain ambient dialogue\cite{mehta2022exploring}. As Game Masters (GMs) in tabletop role-playing games (TTRPGs), LLMs generate plots, characters, and narratives. Applications like AI Dungeon\cite{raley2020playing} use LLMs for interactive storytelling. Tools like CALYPSO\cite{zhu2023calypso} assist human GMs with encounter generation, and Shoelace\cite{acharya2023shoelace} aids in monitoring and responding to in-game conversations.

Additionally, LLMs can narrate game events for players or spectators, enhancing engagement by summarizing interactions and providing automated commentary, which helps streamers manage audience interactions effectively\cite{ranella2023towards}.

Focusing on role-playing games (RPGs), the applications of LLMs have garnered significant attention in both academic research and industry. Existing studies highlight the synergy between LLMs and RPGs. For instance, Generative Agent\cite{park2023generative} introduced computational agents that simulate human behavior and described an architecture that utilizes memories and reflections to dynamic plan agent behaviors. LLMGA\cite{hu2024survey} provides a broad perspective on the architecture and functionality of LLM-based game agents, highlighting their application across various game genres. In study\cite{shanahan2023role}, they examine the nuances of role-playing, particularly the conversational agent's capabilities in deception and self-awareness, providing insights into achieving more human-like interactions in RPGs. Character-LLM\cite{shao2023character} introduces a novel approach to enhancing role-playing scenarios through fine-tuning on role-play datasets, emphasizing the importance of character consistency and improvisation. Rolellm\cite{wang2023rolellm} presents a systematic evaluation of LLMs in role-playing, identifying key areas for improvement and suggesting iterative enhancements based on user feedback. 

These studies have made significant strides in advancing the field, enabling us to explore the transformative potential of AI-native games. Building on this foundation, this paper aims to present a novel perspective on the development and application of AI-native RPGs in the following chapters.

\begin{figure}[t] 
    \centering 
    \includegraphics[width=0.48\textwidth]{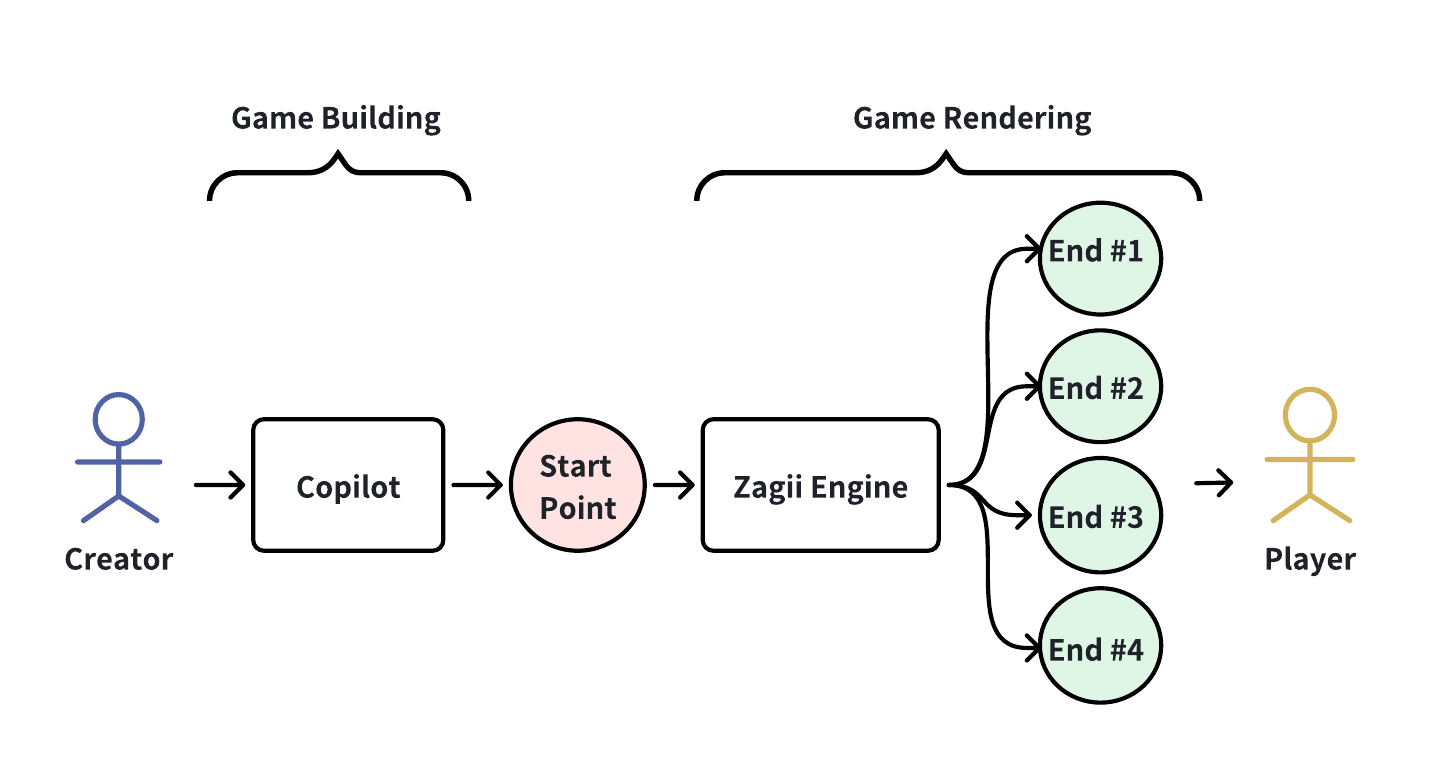} 
    \caption{The text-to-game structure} 
    \label{fig:copilot} 
\end{figure}

The AI-native RPGs we propose share the same fundamental life cycle as traditional games. However, as illustrated in Fig.~\ref{fig:copilot}, each stage is redefined with a fresh approach tailored to the capabilities of AI-driven systems.

Chapter III delves into the details of the Game Building Copilot. Unlike traditional RPG development, which focuses on constructing a complete game, the AI engine primarily establishes the starting framework, including the world setting, characters, and the initial conditions of the game. Chapter IV discusses the Game Rendering process within the AI engine, which surpasses the capabilities of traditional game engines. Acting as the "brain" of the game, it dynamically generates content from the starting framework and adapts in real-time based on user interactions, creating a personalized narrative and ending without human intervention. Chapter V presents the system implementation and experimental results, demonstrating the feasibility of our approach. Finally, this paper concludes by outlining potential key areas for future research and development in this emerging field.

\section{Game Building Copilot}

\begin{figure}[t] 
    \centering 
    \includegraphics[width=0.48\textwidth]{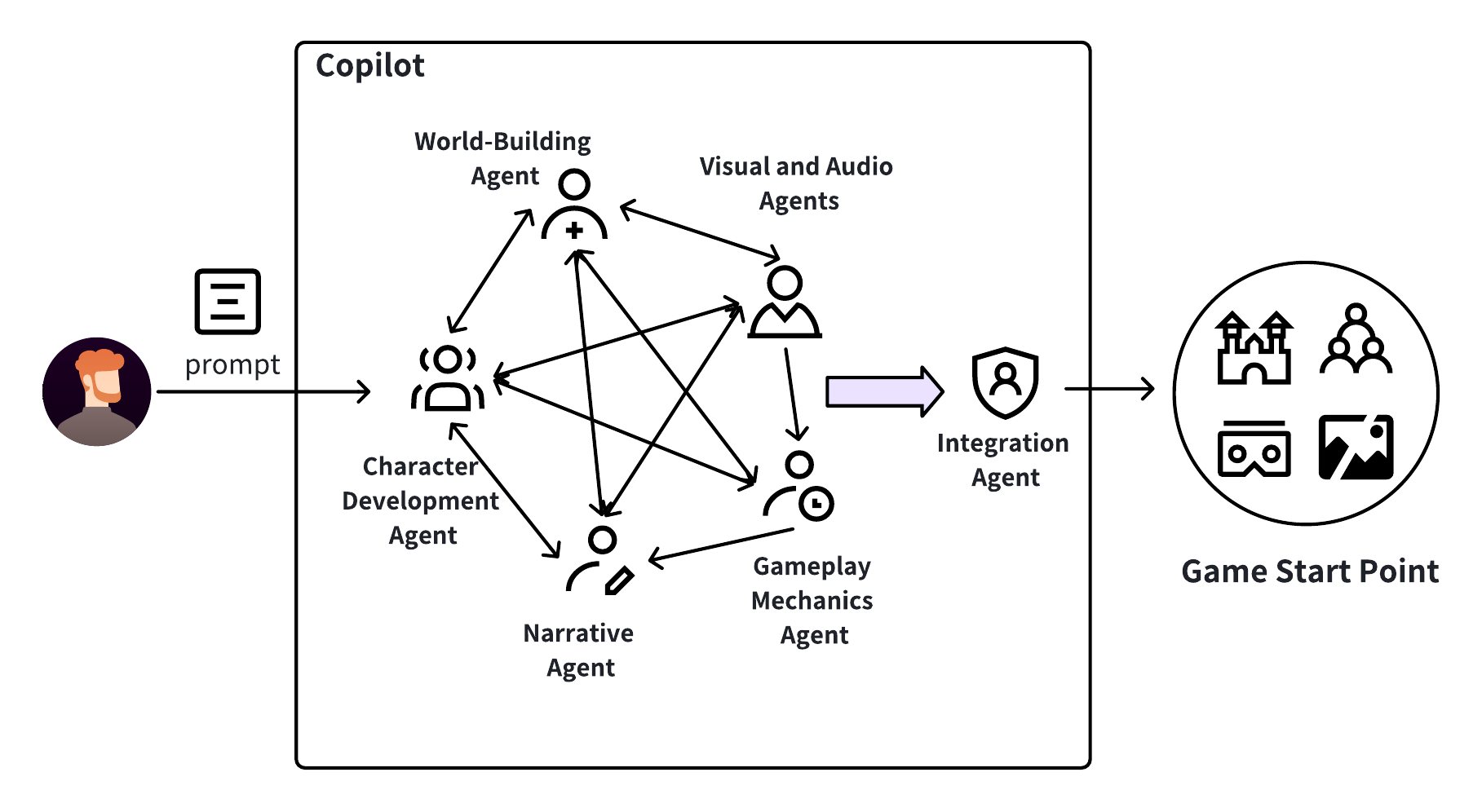} 
    \caption{Multi-agent Game Building Copilot} 
    \label{fig:ma-studio} 
\end{figure}

Imagine the Copilot as a virtual studio composed of multiple AI agents, as illustrated in Fig.~\ref{fig:ma-studio}. Each agent specializes in a specific aspect of game development, working collaboratively to transform a user's brief description into a fully realized game. This multi-agent system expands upon the initial input to generate a comprehensive game setting, including intricate world-building, character creation, and an engaging initial storyline, forming the foundation for an immersive game experience.

A UGC creator can leverage existing IP-based novels to construct familiar worlds or design entirely original settings with unique characteristics. The creator defines the initial state of the world, while the AI engine dynamically collaborates with players throughout the gameplay to shape a personalized and evolving world.

Within this virtual studio, the process begins with the user's input, such as, "A post-apocalyptic world where robots have taken over, and a lone human survivor fights to reclaim their home." The AI agents then collaborate to expand and develop this concept: 

\begin{itemize}
    \item \textbf{World-Building Agent}: Constructs the game's environment, detailing the geography, cities, ruins, and ecosystems of the post-apocalyptic world, creating a vivid backdrop for the narrative.

    \item \textbf{Character Development Agent}: Designs the protagonist, antagonists, and supporting characters, including their backstories, personalities, and motivations, ensuring each character is compelling and integral to the story.

    \item \textbf{Narrative Agent}: Expands the initial plot into a detailed storyline, including key events, conflicts, and resolutions that drive the game's progression. This agent ensures the narrative is engaging and coherent, providing a strong framework for player interactions.

    \item \textbf{Gameplay Mechanics Agent}: Develops the rules and systems that govern player interactions, such as combat mechanics, resource management, and character progression. This ensures that the gameplay is balanced and enjoyable.

    \item \textbf{Visual and Audio Agents}: Generate the visual elements and soundscapes that bring the game world to life, from character models and environment textures to sound effects and music. These agents ensure the game's aesthetic is immersive and cohesive.
 
    \item \textbf{Integration Agent}: Synthesizes the outputs of all other agents, creating a seamless and interactive game experience. This agent ensures harmonious collaboration among all elements, integrating independent settings, and providing guidance for game rendering in Chapter IV to trigger catalysts for player experience during gameplay. This results in a polished and engaging game for the player.
\end{itemize}

The multi-agent virtual studio enables anyone, regardless of technical skill, to initiate the development of complex and immersive games from simple ideas, thereby expanding the creative potential of individual storytellers and small teams. 

The development of a Game Building Copilot is a complex and challenging task, attracting considerable interest from both academia and industry. As it is not the primary focus of this paper, we have provided only a brief overview of the copilot's capabilities and conceptual approach, aiming to help readers better understand the subsequent content. The outputs generated by the copilot constitute the game's meta-information, which will serve as the starting point for the ZAGII game engine to maintain consistency and drive emergent narratives.

\section{AI-Native Game Engine}
As an engine tasked with revolutionizing the next generation game experience, it should embody the following five key characteristics:

\begin{itemize}
    \item \textbf{Zealous}, reflecting the enthusiasm and creativity that the engine brings to game creation.

    \item \textbf{Adaptability}, adjusting and responding to user inputs and preferences.

    \item \textbf{Generativity}, creating content and assets through AI in real-time.

    \item \textbf{Interactivity}, employing real-time, multimedia, and realistic world simulation communication methods to enhance game immersion.

    \item \textbf{Innovation}, as AI is not about mechanically executing tasks, but rather driving the evolution of the game from a god-like perspective.
\end{itemize}

Our “ZAGII” engine serves as the foundational system that drives the game-playing experience, integrating multiple advanced subsystems to create a dynamic and immersive environment while controlling the game progress, for a "Multi-Players, Multi-NPCs" scenario. 

\begin{figure}[t] 
    \centering 
    \includegraphics[width=0.48\textwidth]{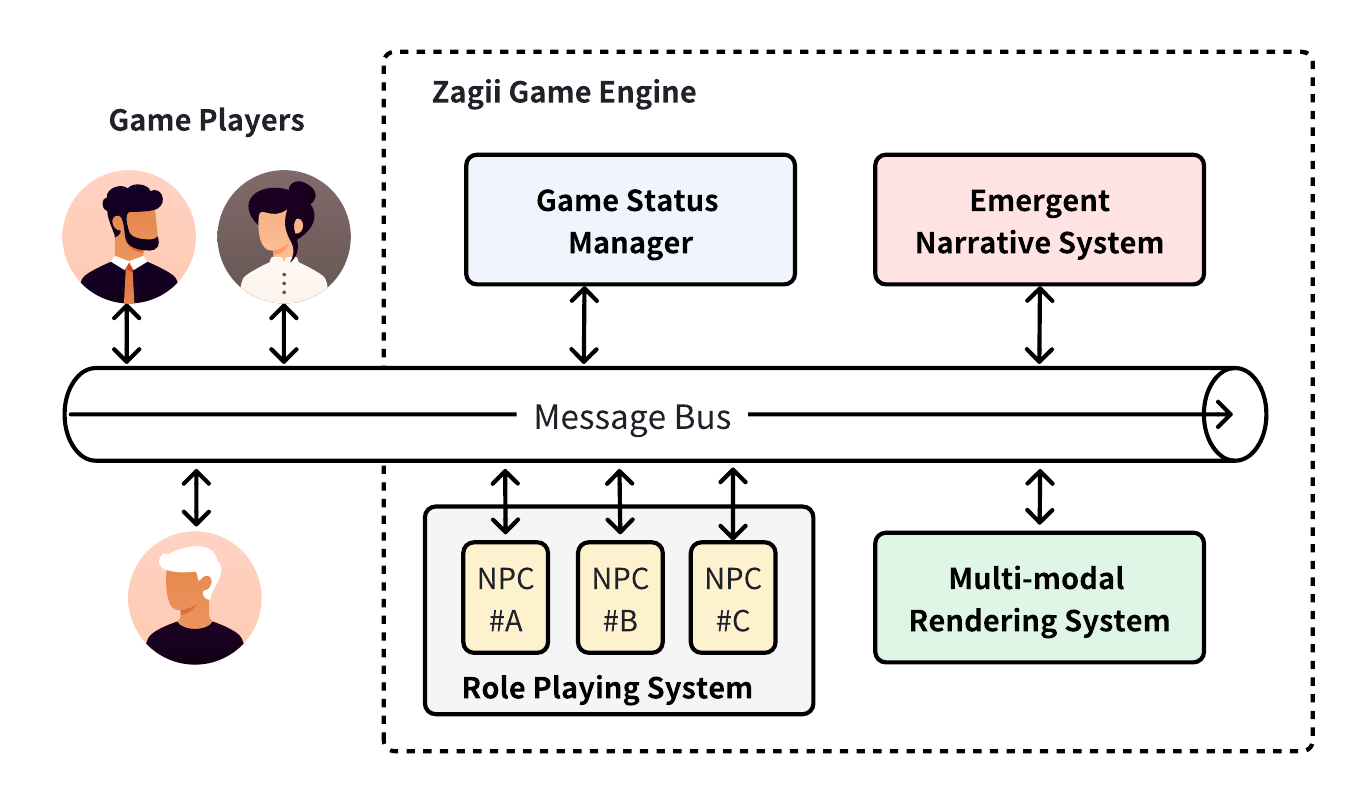} 
    \caption{The conceptual architecture of Zagii Engine} 
    \label{fig:zagii} 
\end{figure}

As shown in Fig.~\ref{fig:zagii}, all modules communicate and exchange information through a centralized Message Bus, ensuring data consistency and system-wide coordination. This integration enables the modules to operate cohesively, resembling a team of agents collaborating toward the shared objective of delivering a seamless gaming experience. By maintaining a consistent flow of information, the Zagii Engine ensures that all game components function in harmony, offering players an innovative and immersive environment.

The Role-playing System leverages the capabilities of Large Language Models to endow NPCs with sophisticated cognitive abilities. These characters can observe their surroundings, understand complex scenarios, think critically, plan their actions, make informed decisions, and interact naturally with their environment. This advanced level of NPC autonomy and intelligence enriches the gameplay experience, making interactions with NPCs more realistic and engaging. By simulating human-like behaviors and responses, the role-playing system creates a more believable and immersive game world.

The Game Status Manager is responsible for real-time tracking and updating various game states, including value attributes, objective environment changes and player’s impact, which is crucial for advancing the game effectively. It monitors the status of game players, NPCs, the environment, and the achievement of game objectives. By continuously updating these states, the Game Status Manager ensures that the game progresses smoothly and that all elements of the game world remain synchronized. This real-time management is essential for creating an immersive and cohesive gaming experience.

The Emergent Narrative System plays a critical role in crafting emergent storylines that both adapts to and catalyzes the ongoing game progress and the decisions made by both players and NPCs as a game designer. Unlike static narratives, these dynamic narratives evolve in response to in-game actions, providing a fresh and original narrative journey for each game session. This adaptability not only encourages players to replay the game due to its endless variability but also ensures that each playthrough offers a unique and engaging experience to emphasize self-impacted fate.

The Multi-modal Rendering System generates sound, music, images, and video based on the current game scenery and progress. Utilizing Diffusion Models and other foundation models, this system creates high-quality game assets that enhance the audiovisual experience. The ability to produce contextually relevant and aesthetically pleasing media on-the-fly adds to the immersive quality of the game, making the virtual world more vibrant and lifelike.

In conclusion, the Zagii Engine's integrated subsystems collectively contribute to a sophisticated and immersive gaming experience. By leveraging advanced AI techniques and real-time data management, the engine ensures dynamic gameplay, responsive interactions, and continuously evolving narratives, thereby setting a new standard for the next generation of interactive entertainment.

In the following chapters, we will discuss the design principles of the Zagii Engine's key subsystems and explain how their implementation was achieved with the assistance of Generative AI.

\subsection{Role-Playing System }

Role-playing Games (RPGs) have become a prominent genre within the gaming industry, offering players a distinctive experience that relies on the interplay between participants and Non-Player Characters (NPCs). These games enable players to immerse themselves in diverse roles, collaboratively shaping a unique narrative through their actions. To elevate this experience, there is a critical need for an intelligent Role-playing System capable of enabling NPCs to authentically perform their roles and actively contribute to advancing the game narrative. Such a system requires NPCs to autonomously generate context-appropriate actions, moving beyond pre-programmed behaviors.

Despite advancements in RPG systems, a major challenge persists: the predominantly reactive nature and constrained response capabilities of current NPCs. While significant progress has been made in developing systems that allow NPCs to respond dynamically to player interactions\cite{urbanek2019learning}\cite{shanahan2023role}\cite{shao2023character}\cite{wang2023rolellm}, empowering these characters to take proactive, goal-driven actions based on their objectives and the evolving game context remains an open-ended and complex problem.

\begin{figure}[b] 
    \centering 
    \includegraphics[width=0.48\textwidth]{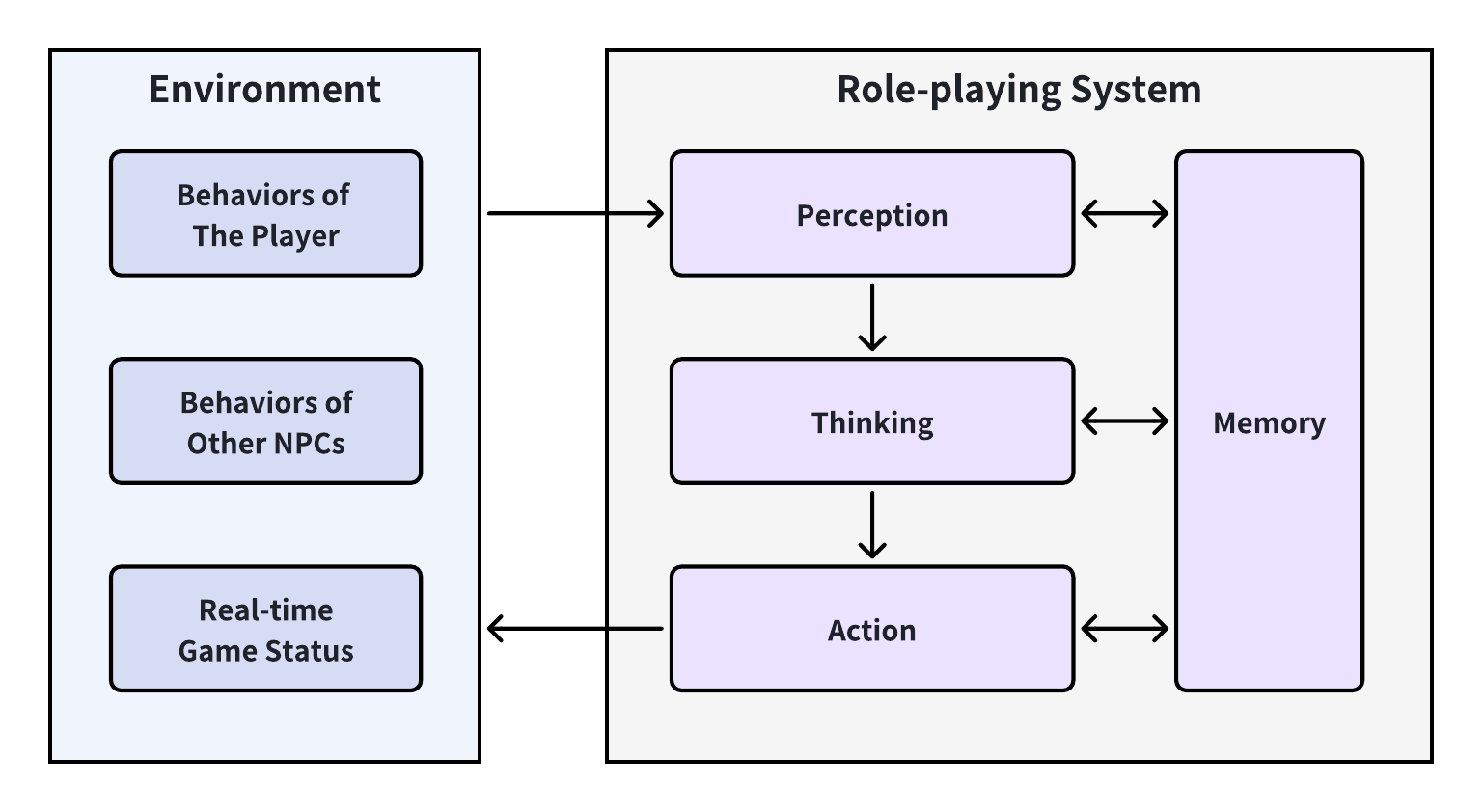} 
    \caption{PMTA framework of Role-playing System} 
    \label{fig:roleplay} 
\end{figure}

To address this limitation, we draw inspiration from the LLM-Powered Autonomous Agents framework\cite{lilian2024ag}, which has demonstrated exceptional potential for human-like decision-making in intricate environments. Building on this foundation, we propose a novel Role-playing framework comprising four core components: Perception, Memory, Thinking, and Action (PMTA), as illustrated in Fig.~\ref{fig:roleplay}. This framework aims to equip NPCs with the ability to perceive their surroundings, retain and utilize contextual information, engage in complex reasoning, and execute appropriate actions autonomously.

The \textbf{Perception} module functions as the character's sensory system, detecting and interpreting all changes within the game world. This includes observing external behaviors, tracking alterations in the game world state, and monitoring the progression of the game narrative. These inputs are processed by the Perception module and converted into structured data that can be utilized by Large Language Models (LLMs) or Multimodal Large Language Models (MLLMs). This processed data serves as the foundation for character decision-making.

The \textbf{Memory} module, another critical component, is responsible for storing essential information relevant to role-playing. This includes the character’s predefined role settings, in-game objectives, self-awareness, and memory fragments generated during role-play interactions. During the decision-making process, the Memory module dynamically retrieves relevant historical memories—represented in natural language text—based on the information currently perceived by the character. This mechanism ensures that characters have access to sufficient contextual information for reasoning and decision-making, thereby maintaining long-term consistency in their behavior. Furthermore, the results of the character's reasoning and actions are carefully analyzed, and any significant information deemed necessary for future interactions is extracted and stored in memory, enabling real-time updates.

The Thinking module processes information received from the Perception module, along with memory fragments retrieved from the Memory module. By employing reasoning, planning, and reflection techniques\cite{lilian2024ag}, it generates action decisions while simultaneously updating the character’s memory. Unlike traditional game AI systems, which primarily rely on methods such as rule-based logic, Finite State Machines, Behavior Trees, Markov Decision Processes (MDP), or Reinforcement Learning for decision-making\cite{uludaugli2023non}, our framework leverages the advanced reasoning capabilities of Large Language Models (LLMs) to handle this complex task. To ensure the Thinking module has access to comprehensive contextual information, we adopt the widely recognized Retrieval-Augmented Generation (RAG)\cite{lewis2020retrieval}\cite{gao2023retrieval} approach. This approach provides essential data, including the character’s role-setting details, memory fragments, real-time game progress, and knowledge about the game world. Additionally, we utilize a dynamic prompt generation module to create personalized role-playing prompt templates for each character. This ensures that characters across various game genres can deliver highly immersive and contextually appropriate role-playing performances.

The Action module interprets and executes the action decisions generated by the Thinking module. These decisions consist of a sequence of action elements, each representing either dialogue or a physical action within the game world. The Action module translates these elements into executable atomic actions within the game engine, rendering them in the game environment. By executing these actions, the module triggers subsequent responses from players or other NPCs and induces changes in the state of objects within the game world.

By integrating the PMTA framework with the reasoning capabilities of LLMs, our system enables characters to autonomously think and make decisions. This not only enhances the quality of the Role-playing System but also has the potential to revolutionize the RPG gaming experience by allowing NPCs to actively participate in and shape the game narrative.

\begin{table*}[t]
\setlength{\extrarowheight}{4pt} 
\caption{The creator's designed objectives are deconstructed and reasoned through from left to right in the table.}
\centering
\begin{tabular}{>{\centering\arraybackslash}m{0.35\textwidth}>{\centering\arraybackslash}m{0.25\textwidth}>{\centering\arraybackslash}m{0.15\textwidth}>{\centering\arraybackslash}m{0.15\textwidth}}
\Xhline{2\arrayrulewidth}
\rowcolor{gray!50} 
\textbf{Goals from creator-written} & \textbf{Subgoals} & \textbf{Anchor of the subgoals} & \textbf{The value of the anchor point at completion} \\ 
\Xhline{2\arrayrulewidth}
\multirow{3}{=}{\textit{\\ Adventurer slayed the evil dragon and escaped from the Black Forest while securing his life and that of the princess.}} 
& The adventurer is safe. & The adventurer's health value & greater than 0 \\ 
\cline{2-4} 
& The princess is safe. & The princess's health value & greater than 0 \\ 
\cline{2-4} 
& The adventurer and the princess escaped from the Black Forest. & Adventurer and Princess Location & Out of the Black Forest \\ 
\Xhline{2\arrayrulewidth}
\end{tabular}
\setlength{\extrarowheight}{0pt} 
\end{table*}

\subsection{Game Status Manager }
The Game Status Manager module in our framework plays a critical role in tracking game progression and enabling the seamless introduction of new plot elements. This module is integral to the gameplay experience, as it determines the appropriate timing for assigning new tasks to players, revealing new clues or plots, and transitioning to subsequent game chapters.

The Game Status Manager is designed to fulfill three key functions:
\begin{itemize}
    \item It analyzes the latest interactions among all game characters, assesses their impact on the game environment, and monitors essential game states. The updated states are then displayed through the user interface (UI), providing players with immediate feedback.
    \item It evaluates whether specific goals or objectives have been achieved based on the current game status.
    \item It identifies when new plot advancements are required by detecting goal completion. In response, the module assigns new tasks to players or NPCs, delivers updated clues or plot information, or concludes the current chapter to initiate the transition to the next one, thereby enhancing the overall playability of the game.
\end{itemize}

During the game development phase, the Game Building Copilot assists creators in identifying critical game statuses that require monitoring. Creators define game goals and establish the criteria for their achievement. The Copilot then determines key performance indicators that align with these goals, enabling continuous tracking throughout gameplay.

Key status details are recorded using numerical values or concise textual descriptions. For example, in emotional companion games, player and character emotions are monitored using intimacy metrics. In Dungeons \& Dragons, the health of players and monsters is tracked, while in adventure games, the player’s current location within the overall map is recorded.

To address complex goals, the Copilot breaks them down into multiple sub-goals based on logical dependencies or relationships. These sub-goals are presented in a structured format to ensure clarity and enable accurate interpretation by the LLMs. An example illustrating this process is provided in Table-I.

The diversity of game goals necessitates a flexible goal-checking module capable of adapting its prompt templates to suit each unique game scenario. This flexibility is crucial, as the module operates continuously throughout game dialogues, requiring both speed and accuracy to ensure seamless gameplay. However, the limited reasoning capabilities of lightweight LLMs can hinder the effectiveness of goal assessments. To mitigate this challenge, we employ two complementary modules that leverage state-of-the-art (SOTA) models:

\begin{itemize}
    \item Cold Start: Before gameplay begins, the SOTA model processes and comprehends the complete scope of the game’s information and objectives. It generates essential considerations for goal validation, which are subsequently incorporated into the goal-checking module's prompt templates. These considerations serve as guidance for the lightweight LLM during gameplay.

    \item Real-Time Assessment: During gameplay, the lightweight LLM’s evaluations are periodically sampled and compared with assessments generated by the SOTA model using identical inputs. This comparative analysis enables the SOTA model to identify and highlight discrepancies or shortcomings in the lightweight LLM’s evaluations, facilitating necessary adjustments to improve accuracy.
\end{itemize}

The Game Status Manager plays a pivotal role in our pursuit of open-ended text-to-game rendering and remains a key focus of our ongoing research.

\subsection{Emergent Narrative System }

Our objective is to revolutionize gaming narratives by creating dynamic, real-time storylines that adapt seamlessly to player actions and the evolving game state. This approach aspires to achieve a "thousand different endings" phenomenon, delivering a distinctive and personalized gaming experience with every playthrough.

Unlike traditional methods that rely on static scripts or predefined storylines, our approach ensures that the gameplay experience remains closely intertwined with the evolving narrative and progression. The Emergent Narrative Generation System is defined by two key features: Real-time Narrative Generation and Interactive Narrative Consumption.

\subsubsection{Real-time Narrative Generation}

Our system generates narratives in real time, aligning story development with game progression and the creator's overarching design. This dynamic generation ensures that the narrative remains both relevant and engaging throughout the gameplay.

One of the primary challenges in game narrative design lies in crafting detailed character interactions and compelling stories. While designers excel at building expansive worlds and intricate frameworks, the creation of nuanced, dynamic narratives often demands additional effort and expertise. This gap has motivated significant research in automated story generation.

Our approach builds upon the principles introduced by Doc\cite{yang2022doc}, which improved long-form story coherence through the use of structured prompts and detailed outlines. Similarly, our system begins with the design of the game world and characters, progressing to the formulation of chapters and goals. Within each chapter, the system enriches the narrative by introducing multiple goals and plot twists, ensuring an engaging and dynamic gameplay experience. Looking ahead, the system will also support the dynamic addition and removal of narrative elements, making it particularly suited for open-world games.

To achieve this, the system leverages contextual information generated during the game-building stage and incorporates a material retrieval mechanism. Specifically, it employs Retrieval-Augmented Generation (RAG) \cite{zhao2024retrieval} to improve the factual accuracy and contextual relevance of the generated narratives.

The workflow integrates the states of players, environments, and NPCs as critical factors in shaping the narrative. During the generation process, data from the game-building stage, current game states, and incomplete goals are synthesized into structured prompts. Additionally, integration with the Game Status Manager ensures that the narrative dynamically adapts to changes in the player's state, environmental conditions, and NPC statuses.

\subsubsection{Interactive Narrative Consumption}

Players primarily engage with the narrative through interactions with NPCs. To facilitate this, our system dynamically updates NPC role-playing prompts to align with the evolving storyline. This ensures that the narrative remains interactive and responsive to player decisions, addressing the limitations of static, pre-defined prompts.

NPC role-playing prompts are structured into three key components: static information, task-related details, and the evolving narrative context. By continuously adapting the narrative context and NPC tasks, the system keeps interactions engaging, relevant, and immersive.

In conclusion, the system enhances game immersion by dynamically generating and presenting narrative elements, maintaining character coherence, supporting open-world dynamics, and seamlessly integrating with the Game Status Manager. This comprehensive approach sustains player engagement by delivering narratives that evolve in response to both player actions and the game environment.

\subsection{Multi-Modal Rendering System }

A complete gaming experience is composed of a combination of multi-modal content including visuals, sound, background music, and sound effects. Building on the foundation of the text-based rendering capabilities, however, unfolding an evolving multi-modal content expression that dynamically responds to player interactions and narrative progress ion throughout the game poses significant challenges on the output consistency and continuous coherent evolution.

The Multi-modal Rendering System utilizes LLMs for memory retrieval, status management, and information orchestration. Through various adapters, it transforms the RPG gaming experience into corresponding multi-modal descriptions to trigger real-time content generation by large multi-modal models. The produced multi-modal content serves as part of the session memory, ensuring consistency throughout the game's evolutionary process. 

\subsubsection{Entities}

Entities are any objects that can act or interact independently, each possessing its own description, attributes, and multi-modal assets, which are part of the session memory and can represent NPCs, scenes, key items, or even players. In AI-native games, when users initiate a game, it merely marks the beginning of an expandable world, initially featuring a limited and incomplete number of entities. Entities are created and updated based on chapter changes and special events within the game. When the multi-modal rendering system renders a scene involving an entity, it reads the entity's multi-modal assets; if these assets are absent, initial assets are generated. If multi-modal assets already exist, they are used as reference information for subsequent generation to maintain consistency. The lifecycle of an entity is determined by the game’s progress, while the lifecycle of its multi-modal assets is determined by that of the entity. 

\subsubsection{Perception and Retrieval}

The perception module is responsible for preprocessing players’ gaming experiences into concise plot summaries, generating information to retrieve entity IDs, and assisting the status manager in updating the status of entity assets. Based on the current round of dialogue, the perception module interprets structured historical dialogue data from the player’s first-person perspective, understanding player intentions and actions, and outputs plot themes and narratives.

Through the interpretation of player behavior by the perception module, the system can retrieve specific entity IDs from session memory storage, representing which entities the user interacted with from their perspective in the current round, and the depth of these interactions. Game Status Manager evaluates changes in entities within the dialogue history based on the current round of dialogue. If there is a significant change in an entity’s textual metadata, its multi-modal assets are updated to reflect the latest game progress. The status of entity assets not retrieved remains unchanged; only the retrieved are used and updated in the current round.

\subsubsection{Adaptive Generation}

Taking image generation as an example, diffusion models guided solely by text struggle to maintain consistency of specific objects across multiple inference processes. Labeled prompts provide a vast representational space, yet they also introduce randomness in the generated content. This inconsistency induced by text-guided conditions poses significant obstacles for rendering real-time RPG games. Reference \cite{yang2024mastering} proposed a training-free framework that uses language models for Recaptioning, Planning, and Generating to guide regional conditional diffusion. The Omost method\cite{lllyasviel2024omost}, proposed by Lvmin Zhang’s team, further summarized the approaches for regional conditional combination diffusion using LLMs, achieving prac tically significant results through uniquely designed block image representation symbols that fine-tune LLMs.

Our multi-modal processor references these studies, implementing regional conditional control of the image canvas through attention decomposition. We first arrange prompts into global and local sub-prompts. For global prompts, corresponding to the canvas background, we use Plot themes and narratives from the perception module as textual prompts to guide the overall image semantics and composition. For local sub-prompts, we employ the Cot method to guide LLMs in regional partitioning of entity locations and introduce image prompts through the IP-Adapter, considering the entity's inherent image assets as reference information to construct image-based regional conditions. By combining multi-modal information and regional conditions in our prompts, our workflow achieves improved semantic accuracy and feature consistency in image generation. The newly generated images are post-processed to update any missing entity assets, thus completing the full cycle of the image generation process.

The generation of sound, music, and motion effects differs in detail from image generation, but the methodology is similar and will not be elaborated further here. 

\section{System Implementation and Experiments  }

\subsection{Implementation of the Experiment System}

This study's innovative nature introduces unique challenges in designing experiments akin to traditional academic research. Specifically, two primary difficulties arise: the lack of publicly available datasets tailored to our research needs and the absence of accessible baseline systems for benchmarking. These limitations have hindered our ability to conduct direct comparisons with related studies, representing a notable constraint of this paper.

\begin{figure}[t] 
    \centering 
    \includegraphics[width=0.48\textwidth]{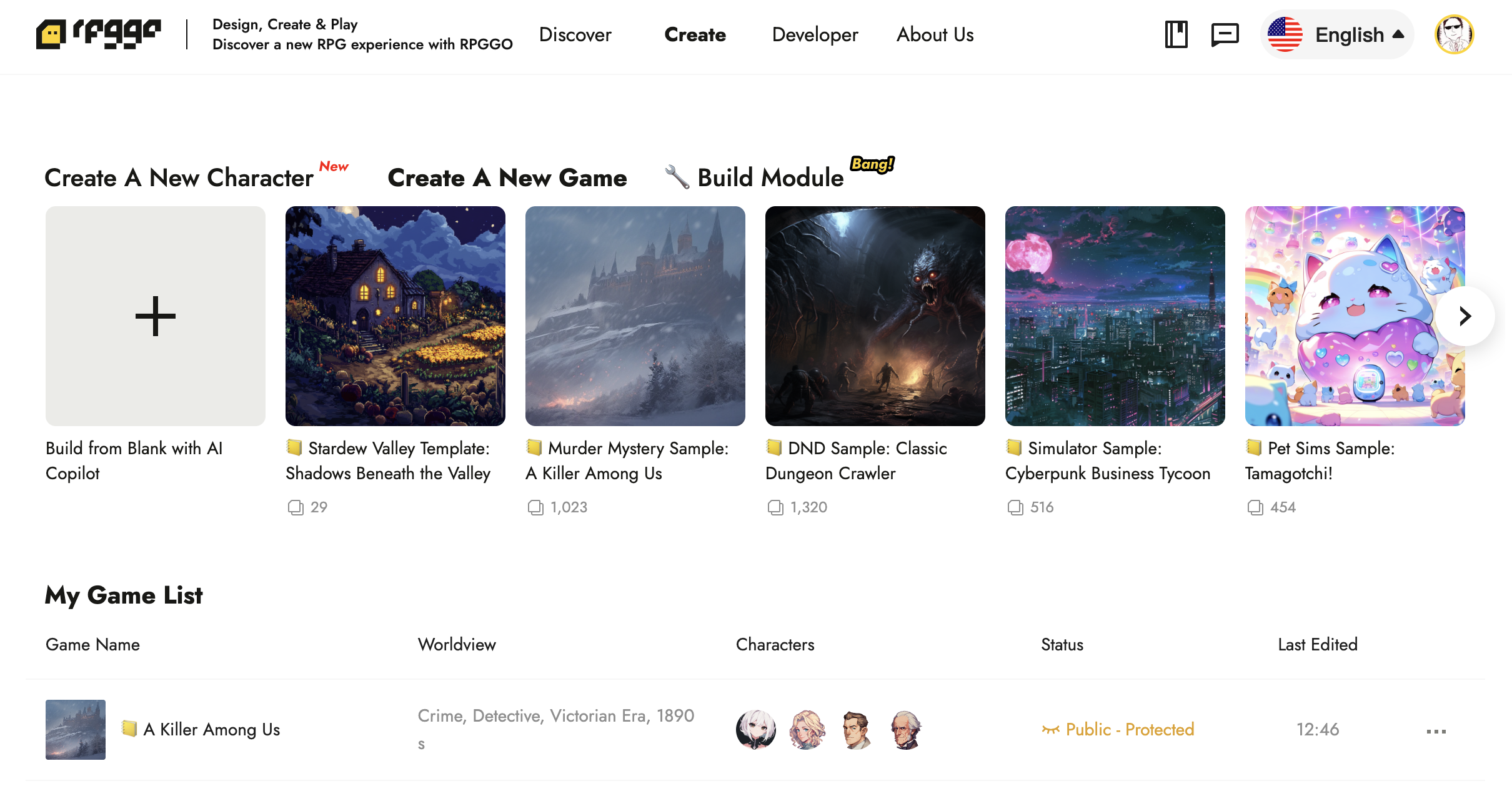} 
    \caption{Game building copilot of our experiment system.} 
    \label{fig:screen-tool} 
\end{figure}

\begin{figure}[t] 
    \centering 
    \includegraphics[width=0.48\textwidth]{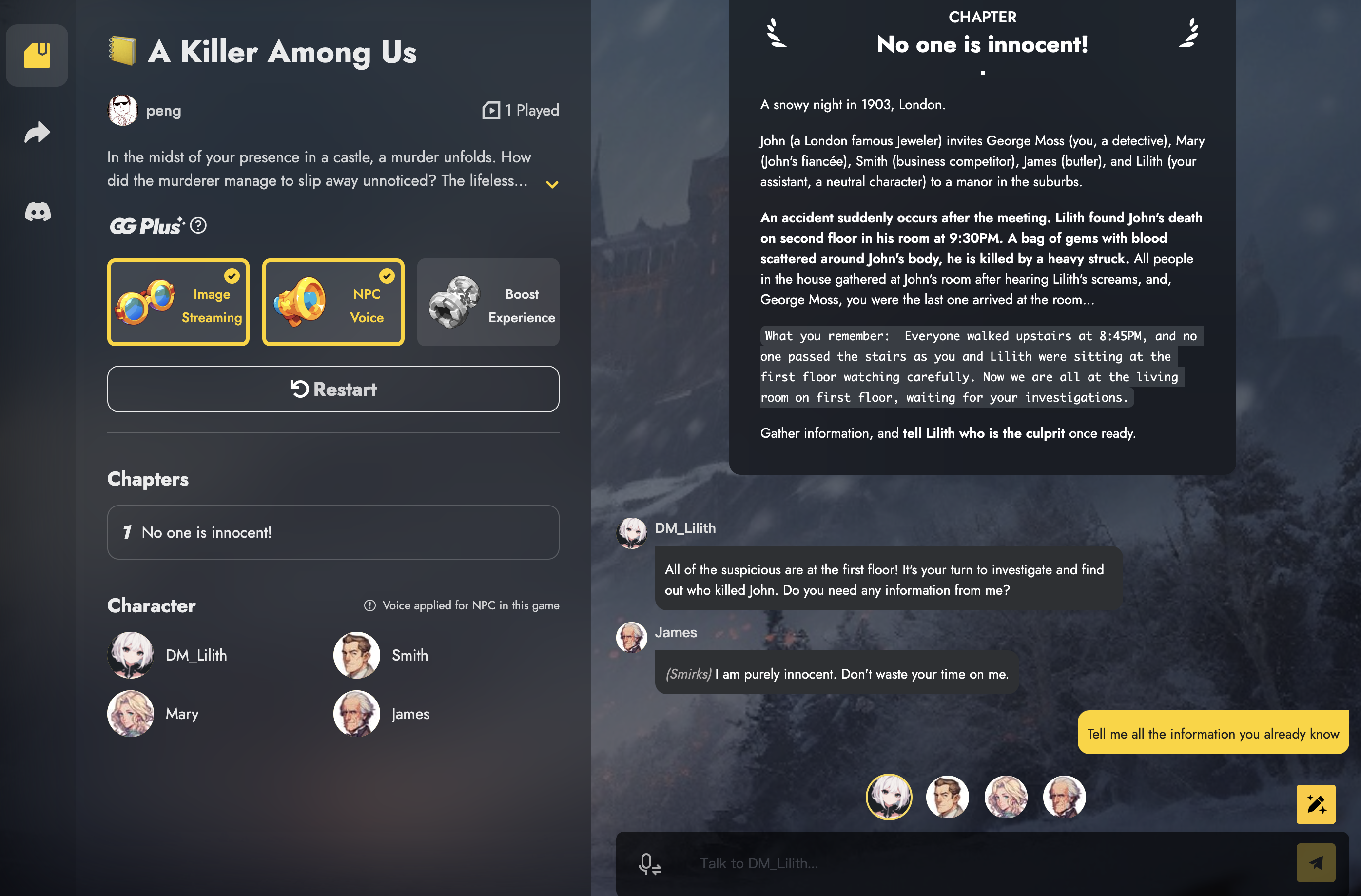} 
    \caption{User can play any published game in our experiment system.} 
    \label{fig:screen-gameplay} 
\end{figure}

To address these challenges, we undertook significant efforts to collect experimental data and validate the proposed solution. An experimental system was developed, integrating the Game Building Copilot and the Zagii Engine, both accessible via our official website\footnote{https://rpggo.ai/}. The Game Building Copilot empowers users to rapidly create games by leveraging the capabilities of large language models, while the Zagii Engine functions as an AI-driven game engine, enabling interactive gameplay responses and multimodal rendering. Screenshots of the experimental system for game building and gameplay are shown in Fig.~\ref{fig:screen-tool} and Fig.~\ref{fig:screen-gameplay}.

 We collected and analyzed user data from our platform over the period spanning February 1, 2024, to May 31, 2024. This dataset includes contributions from 325 game developers and over 60,000 game players. During this testing phase, developers utilized the Game Building Copilot to create a variety of games. Developers were provided with several game templates that could be customized and expanded upon, or they could choose to build entirely original games. With the assistance of the Copilot, most developers published the first version of their games within an hour, enabling rapid playtesting and debugging. The resulting games covered diverse genres, including Adventure, Role-Playing, Mystery, Simulation, and Strategy. Over the study period, a total of 168 games were developed and publicly released, while some developers opted to keep their games private.

\subsection{Experiment Data Analysis }

During the experimental period, the 168 published games collectively recorded a total of 60,301 gameplay sessions. A gameplay session is defined as an instance where a player begins a game and continues until they either complete it or exit midway. Among these, one exceptionally well-designed game accounted for 35,407 gameplay sessions, showcasing its remarkable success and underscoring the potential of the text-to-game framework to create highly engaging content. However, to ensure the generalizability of the data, this outlier was excluded from subsequent analysis. Instead, we focused on the remaining 167 games, which collectively accumulated 24,894 gameplay sessions.

\begin{figure}[t] 
    \centering 
    \includegraphics[width=0.48\textwidth]{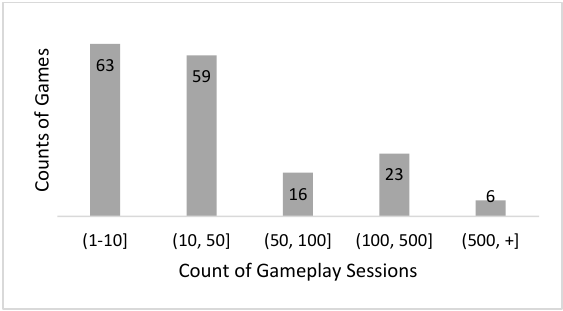} 
    \caption{The distribution of gameplay sessions among the selected 167 games.} 
    \label{fig:gameplay} 
\end{figure}

The distribution of gameplay sessions across these 167 games is illustrated in Fig.~\ref{fig:gameplay}. As shown, a small subset of games attracted the majority of gameplay sessions, aligning with the Pareto principle (80/20 rule) commonly observed in traffic distribution. Specifically, 29 games achieved over 100 gameplay sessions each, while 6 games exceeded 500 sessions.

\begin{figure}[bt] 
    \centering 
    \includegraphics[width=0.48\textwidth]{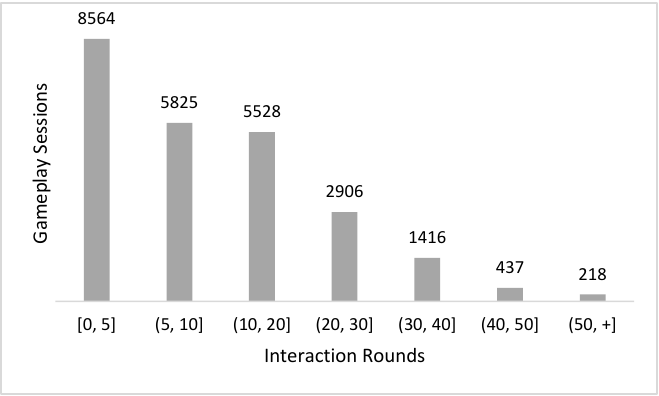} 
    \caption{The distribution of user-npc interaction rounds across all the gameplay sessions.} 
    \label{fig:interaction} 
\end{figure}

Fig.~\ref{fig:interaction} presents the distribution of User-NPC (Non-Player Character) interaction rounds across the 24,894 gameplay sessions. The majority of interaction rounds fell within the range of 5 to 30. Notably, more than 200 gameplay sessions featured interaction rounds exceeding 50. A significant portion of gameplay sessions, however, involved fewer than 5 interaction rounds, reflecting instances where players initiated the game but exited shortly thereafter. These shorter sessions can be attributed to various factors, including technical issues with the game's front-end display or the game failing to meet player expectations.

The experimental data validate the effectiveness of the proposed text-to-game framework: games developed rapidly using the Game Building Copilot and the Zagii Engine demonstrate substantial playability and user engagement. Furthermore, the ability to generate highly popular RPG games was evident. However, challenges such as cold start issues and limited gameplay engagement highlight the need for further optimization of the Zagii Engine and Game Building Copilot to enhance player experience, improve narrative generation, and enable the creation of higher-quality games.

\section{Conclusion and Future Work}
This paper introduces an innovative text-to-game engine capable of creating and rendering RPG games in real time. By leveraging generative AI, the framework simplifies the game development process to a single step—typing in a text box—thereby empowering users to generate their own role-playing games (RPGs). The current implementation supports limited-scale "Single-Player Multi-NPC" RPG scenarios, demonstrating the feasibility of our approach and its potential to transform the game development landscape. As generative AI continues to advance, we foresee the creation of expansive, open-world environments for user-generated content (UGC)-based RPGs, including support for "Multi-Player Multi-NPC" scenarios. This represents the next generation of platforms, akin to Roblox. However, several challenges and limitations must be addressed to fully realize this vision.

To enhance the capabilities of the proposed framework and improve the overall user and player experience, several directions for future work are outlined below:

\subsection{Optimization of the Zagii Engine}
While the Zagii Engine has demonstrated its ability to support real-time gameplay and multi-modal rendering, further refinements are required to enhance its role-playing system, emergent narrative system, game status manager, and multi-modal rendering capabilities. These improvements aim to deliver richer, more immersive gameplay experiences and enable the generation of more dynamic and complex storylines. For instance, refining the emergent narrative system could enable more adaptive and responsive storytelling that reacts to player actions in real time. Similarly, enhancements to the multi-modal rendering system could improve visual fidelity and interactivity, further immersing players in the game world.

\subsection{Refinement of the Game Building Copilot}
The Game Building Copilot has proven effective in simplifying the game creation process, allowing developers to quickly publish playable games. However, further optimization is necessary to improve the quality of the inputs provided to the Zagii Engine. Enhancing the Copilot's ability to interpret user intent and generate high-quality game scripts would significantly elevate the overall quality of the games. Additionally, incorporating advanced natural language processing (NLP) techniques could enable the Copilot to better understand complex user instructions and offer more tailored suggestions for game design.

\subsection{2D \& 3D asset generation}

Generative AI has driven significant advancements in the creation of 2D and 3D game assets. Industry tools such as NVIDIA’s GauGAN and Artbreeder are widely utilized to generate detailed and diverse 2D textures and character designs. In the domain of 3D asset generation, innovations like NVIDIA’s DLSS and Unreal Engine’s MetaHuman Creator have enabled the production of lifelike character models and immersive environments. Academic research has further demonstrated the potential of generative AI to produce high-quality assets that seamlessly integrate into game worlds. Despite these advancements, a key challenge persists: the lack of real-time rendering and optimization algorithms to ensure that generated assets are not only visually impressive but also optimized for performance within game environments.

\subsection{AB Testing Framework}

Another crucial area for future work is the establishment of a robust A/B testing framework tailored for AI-native RPGs. Such a framework would enable large-scale evaluation of different model versions and game features by capturing detailed data points from player interactions. The process involves creating parallel testing environments where players are randomly assigned to different versions of the game engine or specific features. By analyzing player responses and performance across these environments, developers can identify which changes most effectively enhance the gaming experience. To maximize its utility, the A/B testing framework should be both flexible and scalable, allowing for iterative, data-driven improvements to the game engine and features.

\newpage
\bibliographystyle{IEEEtran}
\bibliography{references}

\vspace{12pt}

\end{document}